# Uncertainty Evaluation: Bridging Machine Learning and Measurement Science[1]

Dario Petri, *Fellow IEEE* and Luca Mari [2]

*Abstract* — **Reliable uncertainty evaluation is essential for assessing the trustworthiness of machine learning (ML) predictions, particularly in safety-critical applications. This paper explores conceptual analogies between uncertainty evaluation in ML and measurement science, drawing on the evolution of metrology from error-based approaches to the framework established by the Guide to the Expression of Uncertainty in Measurement (GUM). Particular attention is devoted to the distinction between process-level and result-level quality and to the limitations of the conventional classification of ML uncertainty into aleatoric and epistemic components. It is argued that this dichotomy, similarly to the traditional distinction between random and systematic measurement errors, is inherently model-dependent and provides an insufficient basis for a unified uncertainty evaluation framework. Further analogies between machine learning and measurement field of knowledge are identified concerning distribution shifts, anomalous operating conditions, and the validity of uncertainty evaluation. Building on the proposed considerations, the paper advocates a metrology-inspired approach based on the systematic identification, quantification, and propagation of uncertainty contributions. By extending the methodology already established in measurement science to ML, this approach could offer a potentially rigorous, consistent, and transferable framework for evaluating the trustworthiness of individual predictions.**

## I. INTRODUCTION

In recent years, machine learning (ML) systems have been applied across a wide range of domains involving high-stakes decision-making, including autonomous driving [1], healthcare and medical diagnosis [2]-[4], and climate science [5], [6]. ML systems have also increasingly become integral components of measuring systems [7]-[9]. In all these application fields it is critical to assess and mitigate the risk associated with wrong decisions, and therefore quantifying the trustworthiness of the results of the inference process performed by trained ML systems is an important task, which has become an active and rapidly growing area of research within the ML community [10]-[14]. Possibly also by taking what has been learned on the subject in the context of measurement science, an emerging trend in ML is about modeling trustworthiness in terms of uncertainty: the aim is mainly to support both risk-aware decision making and enhance the explainability of the ML system behavior through the systematic evaluation of the uncertainty of ML predictions[3] [12]. Accordingly, the evaluation of uncertainty is assuming an increasingly important role in a range of different scenarios, including active learning, robustness to data distribution shifts, out-of-distribution detection, and learning under data scarcity [10]-[13]. In these contexts, the ability of a ML system to recognize when to defer a decision, abstain from prediction, or request additional information is often at least as critical as achieving high predictive accuracy.

Once trained, an ML system implements a mapping from input data to inferred class labels or values of a quantity (we do not consider here generative ML systems, which would deserve a different treatment). The trustworthiness of this input-output relationship is affected by several sources of uncertainty, including uncertainties in the input data, model parameters and hyperparameters, limitations inherent in the model architecture, and potential weaknesses in the training procedure. This process exhibits some conceptual analogies with a measurement [15]-[17], in which a set of input quantities is mapped to a measured value, with the resulting output being affected by several sources of uncertainty. These analogies can provide a useful framework for systematically identifying, characterizing, and quantifying the uncertainty contributions that arise in ML models. Indeed, in the context of measurement science the trustworthiness of measurement results has been a widely investigated subject: traditionally addressed in terms of measurement errors, more recent approaches have focused on the evaluation of measurement uncertainty. This evolution is witnessed by the well-established *Guide to the Expression of Uncertainty in Measurement* (GUM) [18], which provides a general framework for evaluating, and expressing measurement uncertainty.

This paper is aimed to draw on the lessons learned on this matter in the field of metrology and to provide insights and, possibly, guidelines for the evaluation of uncertainty in ML systems, hopefully contributing to the establishment of a conceptual link between metrology and ML. Section 2 introduces the general problem of evaluating the quality of a product from the quality of the process by which it is produced, considering both measurement and ML prediction as instances

Dario Petri is with the Department of Industrial Engineering of the Università di Trento, Trento, Italy (e-mail: dario.petri@unitn.it).
Luca Mari is with the Department of Philosophy of the Università Cattolica di Milano, Milano, Italy (e-mail: luca.mari@unicatt.it).

[3] A linguistic note: in ML, the term “prediction” is used independently of a temporal connotation, thus also for past or present events, and for referring to both the process of a trained system and the results it produces; in the present paper the context is sufficient to make the intended meaning clear.

of such processes. Sections 3 and 4 discuss how measurement science has addressed this issue, with emphasis on the evolution from the traditional distinction between random and systematic measurement errors, highlighting the reasons that led to the transition from the error-based approach to one based on measurement uncertainty, as established in the *International Vocabulary of Metrology* (VIM) [19]. Section 5 reviews and critically analyzes the basic criteria adopted for uncertainty modeling and evaluation in ML, highlighting the analogies with the traditional distinction between random and systematic errors. Section 6 discusses the parallels between measurement and ML prediction when operating conditions differ from those expected. Section 7 summarizes the main findings emerging from the comparison between the two bodies of knowledge. Finally, Section 8 proposes some relevant concluding remarks.

## II. A GENERAL PROBLEM: FROM PROCESS QUALITY TO PRODUCT QUALITY?

When a process is designed and operated to generate products – either material or informational –, it may be appropriate to obtain not only the product itself but also information about its quality. This creates a general problem, with both epistemic and operational dimensions: *how can the information about the quality of a process be translated into information about the quality of each individual product generated by that process?*

Indeed, the quality characteristics of a process provide information about the quality of every individual product obtained by that process, but they do not fully determine it. A process can be studied under controlled and repeatable conditions, thereby enabling the characterization of its typical behavior, intrinsic limitations, and sensitivity to relevant influence factors. In contrast, each individual product is obtained under specific operating conditions, and its quality depends both on the process and on the particular input, environment, operating state, and other contingent conditions under which it is produced. Such factors may cause the quality of an individual product to fall below what would be expected from the characteristics of the process. Hence, though exceptions may occur, process quality can generally be regarded as an upper bound on product quality. Information about process quality must therefore be transferred to the individual product and, where appropriate, revised to account for the specific conditions under which the product is generated: nominal or reference characteristics of the process are relevant evidence for assessing the quality of an individual product, but are not, by themselves, sufficient to provide a complete characterization of it.

The case of accuracy is especially instructive in this regard. In general terms, accuracy concerns the degree of agreement between an actual process behavior or product and the corresponding expected or reference process behavior or product. Accuracy thus plays an inverse role with respect to error: accuracy is generally improved by reducing or removing causes of error. Under laboratory or validation conditions, suitable references may be available and the actual products can be carefully observed. This enables the quality of both the process and of its individual products to be reliably evaluated by comparing the observed outcomes with the corresponding reference values or expected outcomes. Under field or deployment conditions, however, the references required to apply such a comparative evaluation to any individual product are usually not available. Consequently, the quality of that product cannot, in general, be assessed using the same reference-based strategy.

This leads us to a distinction that is central to the present analysis:

- *process-level quality*: information about a process, obtained by studying its behavior under a specified set of controlled, calibration, validation, or test conditions;
- *product-level quality*: information about a product, which must be obtained, at least partly, from the specific conditions under which it is actually generated.

Although the two levels are closely related, they are distinct and should not be conflated.

## III. MEASUREMENT AS A PARADIGMATIC CASE

Measurement provides a particularly clear instance of this problem. A measuring system is expected to be calibrated and tested by comparison with suitable measurement standards under controlled conditions. These procedures contribute to set and characterize the behavior of the system and provide information about features such as system sensitivity, selectivity, and stability, which collectively contribute to its accuracy. For a measurement performed under field conditions, however, a reference value for the measurand (ideally its "true value") is usually unavailable. Indeed, establishing such a reference would require a higher-quality measurement of the same measurand, a condition that is seldom fulfilled as it may be technically infeasible or just too costly. Hence, although the notion of accuracy (and the error) of a measurement result remains theoretically meaningful, it cannot generally be operationalized.

For this reason, in the last few decades measurement science has developed a different approach towards the evaluation of the quality of measurement results. Rather than considering the unknown, and usually unknowable, error, this approach focuses on how strongly, and on what evidential basis, a measurement result is warranted by the available prior information, about both the process and the measurand, as well as that acquired through the measurement itself. Taken together, these sources of information contribute to establishing the trustworthiness of the result, without reducing it to a matter of subjective trust. Most probably for continuity with the tradition about measurement error – which is inherently a "negative" concept – such concept of trustworthiness has been called "measurement uncertainty" instead of, for example, "measurement certainty" or "measurement public trust". It is important, however, to emphasize that the concepts of error and uncertainty should not be conflated, as they refer to fundamentally different notions. Indeed, unlike error, measurement uncertainty is currently defined by the VIM as a "non-negative parameter that characterizes the dispersion of quantity values that can reasonably be attributed to the measurand, on the basis of the available information" [19]. Thus, measurement uncertainty quantifies the extent to which the information obtained about the measurand is supported by the available evidence: as such, it is acknowledged to be an essential component of the measurement result.

In summary, the accuracy of a measurement (regarded here as a process-level quality feature) contributes, inversely, to measurement uncertainty (intended as a product-level quality feature). This contribution is sometimes referred to as "instrumental measurement uncertainty". For an individual measurement, performed under specific here-and-now conditions, however, instrumental uncertainty is only one of the several sources contributing to measurement uncertainty, though sometimes dominating.

## IV. The Metrological Transition from Error to Uncertainty

The transition from an error-centered approach to an uncertainty-centered approach did not merely entail a replacement of one numerical indicator with another: rather it involved a change in the conceptual framework used to model and organize the quantification of measurement trustworthiness. This change is characterized in terms of three main structural components of a quality-evaluation framework:

C1. identification of the sources that may affect the quality of the result;

C2. for each source, evaluation of its contribution to the quality of the result;

C3. combination of the individual contributions into an overall indicator characterizing the quality of the result.

The traditional approach focused on C1 by interpreting measurement error as the deviation of the measured value from the true value and by classifying its contributing sources as random errors or systematic errors, according to their behavior. However, this source-behavior based categorization was subsequently abandoned for at least two main reasons. First, when measuring, the true value of the measurand could not be unique, as occurs when the definitional uncertainty is non-null. Moreover, it is usually unknown and in practice unknowable with certainty, making the evaluation of measurement error infeasible. Second, as acknowledged by the GUM (even if the quoted text already considers uncertainty, not error), "such categorization of components of [error] can be ambiguous when generally applied. For example, a 'random' component of [error] in one measurement may become a 'systematic' component of [error] in another measurement in which the result of the first measurement is used as an input datum." [18]. Moreover, distinguishing between 'being random' and 'being systematic' revealed to be a problematic starting point to consistently derive general solutions to C2 and then C3. Indeed, the random-systematic categorization is not an inherent property of an error source, as it may depend on the model adopted to describe the measurement and its context.

Some examples may explain how an error source can exhibit either random or systematic behavior depending on how it enters on the adopted measurement model. Consider an accelerometer whose response depends on the environmental temperature. As long as the influence of temperature on the sensor output is not explicitly modeled, the resulting effect may be considered as a random contribution. Once this dependence is taken into account in the measurement model and the temperature is measured, the corresponding effect becomes systematic and can, at least in principle, be compensated for. The residual effect due to uncertainties in both the compensation relationship and the measured temperature can be classified as a random contribution. As another example, consider an ideal quantizer driven by a pure sine wave. Since both the input signal and the applied transformation are deterministic, the quantization error can, in principle, be determined exactly and thus it exhibits systematic behavior. In practice, however, it is often advantageous to model the quantizer using a simple linear input/output transformation and to treat the error as a random quantity [20].

For these reasons, and following a seminal recommendation of a working group convened in 1980 by the *Bureau International des Poids et Mesures* (BIPM) and then revised and adopted by the *Comité International des Poids et Mesures* (CIPM), the GUM uncertainty framework reverses the strategy. Specifically, it first establishes a requirement for a common quantitative representation of partial indicators (and hence for C2), typically expressed as standard deviations or more generally as probability distributions. Indeed, as stated in [18], "categorizing the methods of evaluating uncertainty components rather than the components themselves avoids [the mentioned] ambiguity. At the same time, it does not preclude collecting individual components that have been evaluated [...] into designated groups to be used for a particular purpose".

The GUM subsequently classifies uncertainty contributions according to the methods used for their evaluation: Type A evaluation uses statistical analysis of series of measured values, whereas Type B evaluation employs methods other than those for Type A evaluation, and then typically relies on prior available information, such as calibration certificates, specifications, previous data, models, or expert knowledge [18]. Accordingly, Type A and Type B do not identify two different kinds of uncertainty, nor do they reproduce the distinction between random and systematic errors: rather, they distinguish the methods used for evaluating uncertainty sources / components.

Hence, within the GUM framework, C1 and C2 lead to the construction of an uncertainty budget, which provides the basis for C3. At this stage, the individual uncertainty contributions are combined through the law of propagation of uncertainty, often based on a first-order Taylor approximation, or by propagating the associated probability distributions, typically by means of Monte Carlo methods. Thus, classifying uncertainty contributions according to the methods used for their evaluation, rather than according to their behaviors, provides an effective and operational means of avoiding the ambiguities discussed above. A further important feature of this framework is that prior information concerning the measurement process is retained and employed into the uncertainty evaluation: calibration results, instrument specifications, and knowledge of typical measurement-system behavior become inputs to the evaluation of the uncertainty for a particular measurement result by integrating them with result-specific information about the actual measurement conditions. Measurement uncertainty can thus be regarded as a result-specific synthesis of heterogeneous, but relevant sources of information.

This does not preclude the identification of contributions according to whether they affect the variability of the measured values in either a random or systematic way, a distinction that remains useful for supporting practical decisions concerning uncertainty-reduction strategies. For example, the effect of contributions exhibiting random behavior can be mitigated through averaging, whereas those exhibiting systematic behavior can be reduced through calibration, compensation, or redesign of the measuring system.

## V. The An analogy between measurement and machine learning inference

The transition from development to deployment of ML systems raises issues that are structurally similar to those encountered in measurement, particularly regarding the requirement of shifting from process-level to product-level quality of information. Indeed, during the development stage, in particular training and validation, the results produced by a ML system can be compared against reference data, analogously to the comparison with reference values performed during the metrological characterization and calibration of a measuring instrument. Performance indexes such as accuracy, precision, sensitivity, and specificity can then be exploited to evaluate the overall quality of the system over a validation dataset, under assumptions concerning the representativeness of that dataset. At deployment time, however, the system produces a result for a new case, precisely because the correct label or value is typically unavailable. Consequently, the accuracy, or the error, of that particular result cannot generally be computed. Although aggregate performance indexes obtained during validation remain informative, they are insufficient to determine the quality of the provided information. Indeed, the quality of a result may be affected by input ambiguity, atypical observations, data acquisition problems, or discrepancies between the operating conditions encountered during deployment and those considered in the training data.

The parallel between calibration and measurement on the one hand and ML training and prediction[4] on the other hand is then manifest:

– under calibration/training and validation conditions, reference values (i.e., values operationally accepted as true labels or true values) are available, so enabling the evaluation of process-level information quality;
– under field/deployment conditions, the reference value associated with each individual result is generally unavailable, so that the quality of the provided information can only be characterized indirectly.

The distinction does not imply that ML lacks established methods for uncertainty evaluation. Approaches such as probabilistic prediction, Bayesian inference, ensemble methods, prediction intervals, conformal prediction, calibration techniques, and out-of-distribution detection already address several important aspects of the problem. Rather, the open issue is whether these methods can be integrated into a more general, process-aware but result-centered framework, analogous to the one developed in measurement science for the modeling, evaluation, and expression of uncertainty. In this context, the conceptual evolution of measurement science from the traditional error-based approach to the current uncertainty-based paradigm may provide useful insights and possibly methodological guidelines for developing a systematic framework for the characterization of the quality of individual ML prediction results.

To quantify and, where possible, reduce predictive uncertainty, considerable attention in the ML literature has been devoted to distinguishing distinct sources of uncertainty according to their underlying nature. This approach is analogous, to some extent, to the traditional classification of random and systematic error contributions adopted in measurement science. In fact, predictive uncertainty is commonly categorized into two main components [21]:

– *aleatoric or data uncertainty*, which refers to the variability of outcomes that remains even when the input-output relationship is adequately known, typically modeled as stochastic variability in the data-generating process;
– *epistemic or model uncertainty*, which arises from incomplete or inadequate knowledge of the underlying input-output relationship; this component may result from limitations in the availability or quality of training data, restrictions imposed by the adopted model class or architecture, deficiencies in the training process, such as underfitting, or discrepancies between the conditions represented in the training data and those encountered during prediction.

The rationale behind this categorization is that it should provide useful guidance to support predictive confidence assessment and model refinement by distinguishing between potentially reducible and inherently irreducible uncertainty sources. Specifically, epistemic uncertainty is considered reducible, at least in principle, through the acquisition of additional training data or the adoption of a more suitable model class. Conversely, aleatoric uncertainty is generally regarded as irreducible because it is assumed to be intrinsic to the considered data [10], [11], thus under the assumption that the inputs to be processed are given, instead of being a sample whose cardinality could be increased.

However, despite its widespread adoption, this categorization has not yet led to the establishment of a unified framework for uncertainty evaluation in ML, and it remains conceptually ambiguous and has been subject to several criticism, including the following:

– analogously to the classical distinction between random and systematic errors in metrology, this categorization is not inherent to uncertainty sources themselves, but it depends on the chosen model, the training procedure, and the information used; this dependence leaves room for arbitrary interpretations and inconsistent applications [14]; for example, the assumption that aleatoric uncertainty is inherently irreducible overlooks the model-dependent nature of its propagation through the system [10], [11];

[4] Note that both training and prediction are inferential processes, of a model and a result respectively.

- the absence of universally accepted definitions of the two types of uncertainty has led to the development of a wide variety of statistical indexes for their quantification [14]; this lack of standardization makes it difficult, if not infeasible, to establish meaningful comparisons among the various uncertainty evaluation methods proposed in the literature, thereby hindering their systematic assessment and further advancement;
- the aleatoric–epistemic categorization may not adequately capture the reducible and irreducible components of prediction uncertainty, highlighting the need for more refined uncertainty classifications that explicitly include the influence of the adopted model and training procedure [14];
- the conventional interpretation of aleatoric uncertainty as inherently irreducible may encourage premature acceptance of existing uncertainty levels, potentially discouraging deeper investigation into their underlying causes and limiting opportunities for model refinement and performance improvement.

Further interesting uncertainty-related analogies between metrology and ML emerge when changes in the system operating conditions occur or when the input data deviates from the assumptions underlying the adopted data-generating model. These aspects are discussed in greater detail in the following Section.

## VI. Analogies between Measurement and Machine Learning Inference under Mismatched Operating Condition

Both measurement and ML processes may be affected by variations in input data, environmental conditions, and/or operating conditions. Such variations may challenge the assumptions underlying system development, potentially compromising, either partially or completely, the validity and reliability of the provided information. Some relevant conceptual analogies between the issues that arise in the two knowledge domains are examined in the following Subsections.

### *A. Dataset or operating condition shift*

In the measurement practice, such situations may arise, for example, from uncompensated systematic effects associated with influence quantities that exceed their specified allowable ranges. Another representative example is the use of a measuring instrument beyond the validity period of its calibration certificate. Under these circumstances, the underlying measurement model may remain conceptually valid, while its accuracy deteriorates because relevant effects are no longer properly accounted for.

A similar situation occurs in ML when prediction is performed on input data whose distribution differs from that of the training set, while the underlying data-generating process remains unchanged [22], [23]. In this case, the input data may still remain within the support of the training distribution, but it could be shifted toward lower data density regions, potentially resulting in higher epistemic uncertainty. Typical examples of distribution shift include language models applied to texts exhibiting writing styles insufficiently represented in the training corpus, or computer vision systems trained primarily on daytime images and subsequently deployed under nighttime conditions. In both cases, the input data remains within the same semantic domain as the training data, and nevertheless the prediction could be biased.

In these scenarios, the assumed measurement or ML model remains fundamentally valid and the information provided is still meaningful. Nevertheless, the associated confidence may decrease due to uncompensated systematic effects in measurement or distribution shift–induced uncertainty in ML.

### *B. Anomalous operating conditions*

Another conceptual analogy between measurement and ML emerges when anomalous inputs (i.e., observations that are inconsistent with the assumed model) occur. In metrology, such observations are commonly associated with *gross errors*, whereas in ML they correspond to *out-of-distribution* (OOD) *inputs*.

In the measurement practice, a gross error typically arises from instrument malfunction, human mistakes, or an incorrect experimental setup. Since these events do not conform to the assumed measurement model, they may invalidate any uncertainty evaluation based on that model. Similarly, in ML, OOD inputs are semantically different from the training data. Unlike distribution shift considered above, OOD data originates from different semantic domains or involve previously unseen classes, thereby violating the assumptions under which the model was trained [22], [23]. As a result, the model may confidently provide predictions that are in fact meaningless, given that the model lacks any information basis to interpret such inputs. As an illustrative example, consider an ML system trained exclusively on sequences of English words: for such a system, most Italian sentences constitute OOD inputs, despite being syntactically valid linguistic expressions, since their characteristics are not adequately represented in the training distribution.

Thus, gross errors in measurement and OOD inputs in ML play analogous conceptual roles: both represent violations of the assumed data-generating model. Consequently, the identification of uncertainty components (e.g., random vs. systematic effects, or aleatoric vs. epistemic sources) as well as the uncertainty evaluation are meaningless. Also, treating OOD inputs or gross errors merely as highly uncertain but valid data (e.g., by modeling them as random effects and averaging them) would lead produce unreliable results that may lead to critical or even catastrophic failures.

It follows that, for ensuring the reliability of the information provided by either measurement or ML systems, a two-step procedure is required [24]:

1. *Model scope verification* (i.e., gross errors or OOD detection): the goal of this step is the identification of anomalous inputs that violate the assumptions underlying the adopted model;

such inputs should be detected and rejected rather than processed using a model that is not applicable to them;

2. *System output and uncertainty assessment*: only after verifying that the input data is consistent with the assumptions and validity domain of the adopted model, the system output should be provided and its uncertainty evaluated; this ensures that the resulting uncertainty is meaningfully evaluated.

Finally, it is worth observing that gross errors in measurement may indicate not only instrument or procedural failures, but also previously unrecognized physical phenomena, potentially motivating the development of new measurement models or new theoretical frameworks. Analogously, while some OOD inputs should be rejected, others may represent novel but legitimate data, thereby motivating model extension or retraining. In both domains, the decision to reject anomalous observations or incorporate them through model updating depends on the available prior knowledge, the underlying causes of the observed deviations and the specific application context.

## VII. Summary

The preceding analysis about the different approaches employed in metrology and ML for uncertainty evaluation is summarized in Table I, which provides a comparative overview of their main characteristics.

Table I. Comparison of the main characteristics of uncertainty evaluation approaches adopted in metrology and machine learning.

| **classification criterion** | **measurement** | **machine learning** |
|---|---|---|
| uncertainty classes | statistical (type A) evaluation; other than statistical (type B) evaluation | aleatoric or data uncertainty; epistemic or model uncertainty |
| theoretical framework | probability theory | probability theory |
| criterion motivated by … | evaluation | evaluation, reduction |
| result depends on … | adopted measurement model, measuring instruments quality, calibration process | adopted data model and training process |
| uncertainty representation approach | standard uncertainty, expanded uncertainty, pdf | not univocally defined |
| combination of uncertainty components | standardized rules based on probability theory | not univocally defined |

The main findings concerning the conceptual analogies between the behavior of measurements and ML predictions under operating conditions that violate the assumptions established during system development are summarized in Table II.

Table II. Comparison of the behavior of measurements and machine learning inferences under operating conditions that violate the assumptions underlying the respective adopted models.

| **operating conditions mismatch** | **measurement** | **machine learning** |
|---|---|---|
| dataset or operating conditions shift | uncompensated systematic effects, increased uncertainty | distribution shift, increased epistemic uncertainty |
| anomalous operating conditions | gross error, rejected input | out-of-distribution data, rejected input |

## VIII. Conclusions

Limitations of the uncertainty evaluation approach currently applied in ML can be more clearly interpreted when examined through the lens of the metrological framework for uncertainty evaluation. In particular, the widespread aleatoric–epistemic dichotomy exhibits strong conceptual similarities to the traditional distinction between random and systematic errors employed in the pre-GUM metrology framework. The main weakness of that distinction is that it is not intrinsic to the uncertainty sources, but rather it depends (either implicitly or explicitly) on the adopted model, the training procedure, and the information used. Consequently, individual uncertainty contributions may shift from one class to the other depending on changes in the model structure, training methodology, or information available, in direct analogy with the context-dependent interpretation of random and systematic errors in measurement science.

This situation parallels the one that motivated the GUM to move beyond the traditional error-based classification in favor of a unified probabilistic framework for uncertainty modeling, evaluation, and expression grounded in the kind of use of the available information rather than the presumed nature of uncertainty sources. In addition, the coexistence in ML of multiple, non-equivalent statistical description for uncertainty modeling and evaluation contrasts with the standardized and internally consistent methodology established by the GUM. Within this framework, all recognized sources of uncertainty, regardless of their origin, are represented through appropriate probability distributions, conditional on the specified measurement conditions, and propagated through the measurement model to achieve the combined uncertainty of the measurement result.

Finally, the common ML interpretation of aleatoric uncertainty as irreducible and epistemic uncertainty as reducible may be seen as an oversimplification when considered from a metrological perspective. In metrology, uncertainty reduction is not simply based on categorical classification, but on a systematic analysis of individual uncertainty contributions and their propagation through the measurement model. This approach enables the identification and quantification of the relevant contributions, as well as the assessment of potential strategies for their mitigation, including improved modeling, refined calibration procedures, and/or measuring system redesign.

In conclusion, we argue that shifting beyond the binary classification of aleatoric and epistemic uncertainty toward a GUM-inspired framework, grounded on the systematic identification, quantification and propagation of all recognized uncertainty contributions, could provide a more rigorous and consistent basis for uncertainty evaluation in ML. Furthermore, from this framework ML could draw some pragmatic guidelines about the identification of the uncertainty boundaries for an appropriate information-enabled decision-making, as modeled in metrology in terms of definitional uncertainty and target uncertainty [19], [25], [26]. Such an approach could promote the development of more reliable, transparent and transferable uncertainty evaluation methods, while supporting a more systematic assessment of opportunities for uncertainty reduction.